\documentclass[sigconf]{aamas}

\usepackage{balance} 

\usepackage{framed}

\usepackage{pifont}

\usepackage{hyperref}

\usepackage{pdfpages}

\usepackage{todonotes}

\usepackage[belowskip=-15pt,aboveskip=0pt]{caption}

\makeatletter
\gdef\@copyrightpermission{
  \begin{minipage}{0.2\columnwidth}
   \href{https://creativecommons.org/licenses/by/4.0/}{\includegraphics[width=0.90\textwidth]{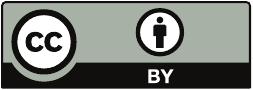}}
  \end{minipage}\hfill
  \begin{minipage}{0.8\columnwidth}
   \href{https://creativecommons.org/licenses/by/4.0/}{This work is licensed under a Creative Commons Attribution International 4.0 License.}
  \end{minipage}
  \vspace{5pt}
}
\makeatother

\setcopyright{ifaamas}
\acmConference[AAMAS '25]{Proc.\@ of the 24th International Conference
on Autonomous Agents and Multiagent Systems (AAMAS 2025)}{May 19 -- 23, 2025}
{Detroit, Michigan, USA}{Y.~Vorobeychik, S.~Das, A.~Nowé  (eds.)}
\copyrightyear{2025}
\acmYear{2025}
\acmDOI{}
\acmPrice{}
\acmISBN{}

\acmSubmissionID{367} 

\title{A Scoresheet for Explainable AI}

\author{\href{https://orcid.org/0000-0002-5545-7003}{Michael Winikoff}}
\affiliation{
  \institution{Victoria University of Wellington}
  \city{Welington}
  \country{New Zealand}}
\email{michael.winikoff@vuw.ac.nz}

\author{\href{https://orcid.org/0000-0002-7699-6444}{John Thangarajah}}
\affiliation{
  \institution{RMIT University}
  \city{Melbourne}
  \country{Australia}}
\email{john.thangarajah@rmit.edu.au}

\author{\href{https://orcid.org/0000-0002-0514-9221}{Sebastian Rodriguez}}
\affiliation{
  \institution{RMIT University}
  \city{Melbourne}
  \country{Australia}}
\email{sebastian.rodriguez@rmit.edu.au}

\begin{abstract}
Explainability is important for the transparency of autonomous and intelligent systems and for helping to support the development of appropriate levels of trust. There has been considerable work on developing approaches for explaining systems and there are standards that specify requirements for transparency. However, there is a gap: the standards are too high-level and do not adequately specify requirements for \emph{explainability}. This paper develops a scoresheet that can be used to specify explainability requirements or to assess the explainability aspects provided for particular applications. The scoresheet is developed by considering the requirements of a range of stakeholders and is applicable to Multiagent Systems as well as other AI technologies. We also provide guidance for how to use the scoresheet and illustrate its generality and usefulness by applying it to a range of applications.
\end{abstract}


\newcommand{\BibTeX}{\rm B\kern-.05em{\sc i\kern-.025em b}\kern-.08em\TeX}

\newcommand{\cb}{$\Box$}
\newcommand{\formline}{\rule{\linewidth}{0.5pt}}

\newcommand{\sectiondivider}{\centerline{\rule{\textwidth}{0.5pt}}}
\renewcommand{\sectiondivider}{\centerline{\textbf{\LARGE * * *} }} 

\begin{document}


\pagestyle{fancy}
\fancyhead{}


\maketitle 


\section{Introduction}\label{sec:intro}

It is important for autonomous and intelligent systems\footnote{Terminology: Since we consider both autonomous systems and other systems that use a range of Artificial Intelligence techniques, we use ``autonomous and intelligent systems'', sometimes compressed to just ``intelligent systems''. We also avoid the term ``model'' (unless we are specifically talking about machine learning) in favour of ``module''.  Finally, we use ``behaviour'' as shorthand for ``behaviour or outcome'' which encompasses the system taking action or providing some output (e.g.~a classification or recommendation).} to be explainable for a range of reasons. Providing explanations can be required by legislation either directly (e.g.~GDPR\footnote{\url{https://data.consilium.europa.eu/doc/document/ST-5419-2016-INIT/en/pdf}}) or indirectly as a consequence of legislation~\cite{DBLP:journals/internet/WinikoffSSM21}.
Providing explanations can also play a crucial role in helping to make autonomous and intelligent systems socially acceptable~\cite{Floridi2018}, transparent~\cite{DBLP:journals/firai/WinfieldBDEHJMO21,DBLP:conf/atal/AnjomshoaeNCF19}, understandable~\cite{DBLP:conf/atal/VerhagenNT21}, accountable~\cite{DBLP:conf/at/CranefieldOV18}, and to help establish an appropriate level of trust~\cite{DBLP:conf/aaai/LangleyMSC17,EMAS2017Winikoff,DBLP:conf/hri/RobinetteLAHW16,Floridi2018,DBLP:conf/atal/SRJT24a,DBLP:conf/atal/SRJT24b}. 

The importance of explainability has also been recognised by various standards. 
For instance, the Ethics Guidelines for Trustworthy AI\footnote{\url{https://digital-strategy.ec.europa.eu/en/library/ethics-guidelines-trustworthy-ai}} and subsequent Assessment List for Trustworthy Artificial Intelligence (ALTAI) for self-assessment\footnote{\url{https://digital-strategy.ec.europa.eu/en/library/assessment-list-trustworthy-artificial-intelligence-altai-self-assessment}} consider explainability as one of a range of factors (e.g.~human agency and oversight, accountability, societal \& environmental well-being). IEEE P7001~\cite{IEEEP7001} also considers explainability as part of transparency, and defines a number of requirements relating to explainability (e.g.~that information is provided on how a system works in general, or that the system provides the ability to answer ``why?'' questions). 

However, this work does not provide adequate guidance for the development and evaluation of the explainability of systems. The Ethics Guidelines for Trustworthy AI only poses questions that ask whether the decisions and outcomes can be understood and whether an explanation is provided, and the Assessment List  has just two questions: ``Did you explain the decision(s) of the AI system to the users?'' and ``Do you continuously survey the users if they understand the decision(s) of the AI system?''. Similarly, IEEE P7001 only provides a few explainability requirements (``why?'' and ``what if?'' questions, as well as global\footnote{A common distinction \cite{DBLP:journals/fcomp/HoffmanMKJT23,DBLP:conf/chi/LiaoGM20} is between \emph{local} explanations that relate to a specific execution (e.g.~``why did you do this?''), and \emph{global} explanations that are not about a specific execution, and hence more general, but necessarily less detailed.} explanation - see~\S\ref{sec:relatedwork}), and Hoffman \textit{et al.}~\cite{DBLP:journals/fcomp/HoffmanJTKM23} assign each system only a single number (1-7).

Following IEEE P7001, we propose to provide this guidance in the form of a scoresheet\footnote{Our scoresheet does not use numbers, but it contains more than just checkboxes, so we use the term ``scoresheet'' for consistency with P7001.}. The P7001 scoresheet focuses on transparency, and is complementary to our scoresheet: our scoresheet is specifically for explainability, and provides details, whereas P7001 has considerably less detail on explainability (see~\S\ref{sec:compare7001}).

The scoresheet can be used in various ways with the most obvious being to evaluate the explainability of candidate systems. 
Used this way, the responses affect which system is chosen because the scoresheet captures that a crucial explainability aspect is lacking, or that another system provides it better.

Explanations are used by different people for different purposes~\cite{DBLP:journals/misq/GregorB99,Biran2014,DBLP:conf/fat/MittelstadtRW19,DBLP:journals/ai/LangerOSHKSSB21,DBLP:conf/re/BuitenDS23}, and therefore we develop our scoresheet by considering the explainability needs of different stakeholders (\S\ref{sec:relatedwork}). 

This paper makes a number of contributions.
Firstly, we develop (and justify) a scoresheet\footnote{The scoresheet was developed iteratively (define, apply, revise); the version presented in \S\ref{sec:scoresheet} is the final one.} for explainability (\S\ref{sec:scoresheet}). 
Secondly, we provide additional detailed guidance on \emph{how} to complete the scoresheet (\S\ref{sec:operationalise}), including an additional checklist for global explanations. 
Thirdly, we demonstrate that the scoresheet is applicable to a range of systems (\S\ref{sec:apply}), showing that the scoresheet is \emph{usable} and \emph{generic}, as well as that it is \emph{useful} (i.e.~that it provides a useful summary).

\section{Stakeholder Explainability Needs}\label{sec:relatedwork}

IEEE P7001~\cite{DBLP:journals/firai/WinfieldBDEHJMO21,IEEEP7001} defines five stakeholder groups: end users, wider public \& bystanders, safety certifiers, incident/accident investigators, and lawyers \& expert witnesses. They consider the range of forms of transparency that each requires. For instance, that end users might want to be able to get natural language answers to ``why did you do that?'' and ``what would you do if \ldots?'' questions. Or that safety certifiers need information on what steps were taken to verify and validate a system. They go on to propose a simple transparency scale for each of the five stakeholder groups. For example, for an end user, the levels can be summarised as: 0: ``no transparency''; 1: information provided on how the system works in general (including, if relevant, on data used); 2: same as 1, but interactive; 3: ability to answer``why?'' questions for specific cases; 4: ability to answer hypothetical ``what if?'' questions; and 5: provision of ``continuous explanation \ldots that  adapts \ldots based on the user's information needs and context''. 

Arya \textit{et al.}~\cite{DBLP:journals/corr/abs-1909-03012} argue that different stakeholders require different sorts of explanations. They propose a taxonomy (and associated toolkit) that allows stakeholders to select an explanation method that suits their needs. Their context is narrower than ours (machine learning systems that learn from data). Their taxonomy considers factors such as the following. Are explanations (of data) given as particular features (e.g.~income or level of debt), examples, or distributions? Do explanations explain individual cases or overall behaviour (local vs.~global)? Is the explanation derived directly from the model used to make decisions, or from another (surrogate) model? 
Elements of their taxonomy are relevant to our scoresheet, and are incorporated in Section~\ref{sec:scoresheet}. These are:
explanation of data (where relevant, using examples, distributions, features) vs. explanation of the model/module;
the distinction between global and local explanations (which is also raised by other literature); and the distinction between an explanation being derived from the module itself, or from a surrogate\footnote{Their taxonomy has two versions of this: for local explanations they distinguish between a self-explaining model and post-hoc explanations, whereas for global explanations they distinguish between directly interpretable models and post-hoc explanations such as a surrogate model, or a visualisation.}. 

Liao \textit{et al.}~\cite{DBLP:conf/chi/LiaoGM20} interviewed 20 UX and design practitioners from IBM to ``identify gaps between the current XAI [eXplainable AI] algorithmic work and practices to create explainable AI products''. Their focus is narrower than ours (explanations of machine learning for end users). 
One useful contribution of their work is their interview framework: they developed a bank of questions to ensure that the interviews covered a range of important aspects. In order to develop this, they identified a range of question types that can be addressed by current XAI methods, including both widely used questions (How, Why, Why not, What if) and less widely-used questions (how to be that, how to still be this; explained in Section~\ref{sec:scoresheet}). Their XAI question bank covered six topics: input (i.e.~data used), outputs produced, performance (e.g.~accuracy, precision, limitations), how (global), why \& why-not (one topic), and a topic covering hypothetical questions (what if, how to be that, how to still be this). 

The most directly relevant work to establishing stakeholder needs for explainability is the recent paper by Hoffman \textit{et al.}~\cite{DBLP:journals/fcomp/HoffmanMKJT23} which seeks to establish what various stakeholders need by interviewing a range of stakeholders. One key point that they identify in their interviews is that the assumption that there are distinct, clearly distinguishable, stakeholders does not necessarily hold. Rather, they found that people had different roles, but that they adopted the viewpoints of different roles at different times, including roles other than their own.  They highlighted the need for both global explanations (that are not too high-level, including holistic performance aspects such as biases, assumptions, bounding conditions and limitations) and local ones, and noted that it can be desirable to link them by having global explanations that refer to particular cases. They flagged the particular importance of edge cases in understanding how the system operates, and what are its limitations.
More broadly, they identified the benefit of having access to the system development team and to (trusted) domain practitioners, and of having information about the system's context (e.g. what does it integrate with, how does it support users' goals) and the role of the company making the software, and trust in it, in a broader accountability and responsibility context.

\section{An XAI scoresheet}\label{sec:scoresheet}

In this section we present the XAI scoresheet, focusing on \emph{what} is included, and \emph{why} it is included.  Section~\ref{sec:operationalise} provides guidance on \emph{how} to use the scoresheet.

The XAI scoresheet (Figure~\ref{fig:scoresheet}) has a number of sections that each collect different information. An initial section collects some \textbf{basic information}. Then there is a section that focuses on \textbf{veracity}, then \textbf{global explanations}, and finally a section focusing on a range of information relating to \textbf{local explanations}: features of explanations, the concepts used, the explanation types supported, and the level of automation. 

\textbf{Basic information:} There are two pieces of basic information that the scoresheet collects. Firstly, whether the system's source code and (if relevant) training data is available. This is useful to know because access to code (and data) can help in understanding explanations, and in assessing the system's veracity (see below). However, this is of more use if there is access to the developers of the system, who can help to navigate the code (and data), and to (trusted) domain experts who can help to explain the context of use. Hoffman \textit{et al.}~\cite{DBLP:journals/fcomp/HoffmanMKJT23} found that access to the system's developers and to trusted domain experts can be important to help understand the system's operation. In the case where the organisation assessing or using the system is also the one that is developing the system, then both these criteria would normally be met.

\textbf{Veracity:} An important basic requirement of explanations is that they actually correspond to the system's reasoning. An explanation system that invents explanations that do not reflect the actual reasons is clearly not useful, and could in fact mislead, and therefore be worse than not having an explanation at all. We therefore include in the scoresheet a high-level question to indicate the reliability\footnote{We use ``reliability'' in the Cambridge dictionary sense of ``the quality of being able to be trusted or believed because of working or behaving well''.} of  explanations  (Low/High\footnote{If the system does not provide local explanations then veracity is not applicable.}). 

One approach to providing explanations with high reliability is to generate explanations directly from either the actual module used to make decisions, or from a log that records what the system actually did and the factors considered (termed a ``blackbox''\footnote{As in an aircraft blackbox}~\cite{DBLP:journals/firai/WinfieldBDEHJMO21,WinikoffEMAS24}). This direct approach provides a high level of confidence that the explanation reflects the actual reasons. 

An alternative approach is to construct explanations using an alternative proxy model. In this case it is possible for explanations to not correspond to the actual reasons, and so steps need to be taken to attempt to ensure alignment between the behaviour-generating model and the explanation model, and to assess the effectiveness of the alignment. 
For example, alignment can be attempted to be ensured by deriving the proxy model from the actual model by a systematic process or algorithm, and the alignment can be assessed by having a process of testing that evaluates for a range of system behaviours and explanations whether the generated explanation matches up with the real reasons for the system's behaviour. These real reasons can be identified by running the system on hypothetical scenarios to confirm that varying the reasons results in a change in behaviour. In some cases they may also be able to be identified by adding debugging probes to the system.

Regardless of the approach taken, it is important in order to be able to trust the explanations to know not just that explanations are reliable, but also \emph{why} they are reliable, and so the scoresheet captures this information.

\begin{figure}
\begin{framed}
\begin{tabbing}
    XX \= XXX \= XXXX \= XXX \= \kill

    \centerline{\textbf{XAI scoresheet for} \rule{2cm}{0.5pt}} \\
    \cb\ System source code is available \\ 
    Is training data used available? Yes / No / Not Applicable \\
    \cb\ There is access to the system's developers \\
    \cb\ There is access to trusted domain experts \\[1.5mm] 
    
    \textbf{Veracity:} \\
    How reliable are explanations? Not Applicable / Low / High \\
    What steps are taken to ensure explanation reliability? \\
    \formline \\
    \sectiondivider \\
    \textbf{Global Explanations}: Has information been provided on: \\ 
        \> \cb\ \emph{How} does the system work? \\ 
        \> \cb\ \emph{How well} does it work? \\
        \centerline{\textit{(See checklist - Figure~\ref{fig:checklist})}} \\ 
 \sectiondivider\\
 \textbf{Local Explanations}: 
     Explanations \ldots \\
     \> \cb\ \ldots can be \textbf{individually customised} \\ 
     \> \cb\ \ldots are \textbf{interactive} \\
     \> \cb\ \ldots include an indication of \textbf{confidence} \\
     \> \cb\ \ldots include an indication of \textbf{scope of generalisation} \\[1.5mm]
What \textbf{Concepts} are used in explanations? \\ 
    \> \cb\ Examples \cb\ Features \cb\ Beliefs \cb\ Events/Percepts \\
    \> \cb\ Goals  \cb\ Actions \cb\ Preferences \cb\ Values \\
    \> \cb\ Other: \> \>  \rule{5.5cm}{0.5pt} \\[1.5mm] 
    What forms of \textbf{Explanation Types} are provided? \\
    Factual/Past: \> \> \> \cb\ Did? \cb\ Why? \cb\ Why not? \cb\ Contrastive \\ 
    Future-looking: \> \> \> \cb\ Will? \cb\ Why? \cb\ Why not? \cb\ Contrastive \\ 
       Hypothetical: \> \> \> \cb\ What-if? \cb\ How to be? \cb\ How to still be?
       \\
       Other: \> \> \> \rule{5.5cm}{0.5pt} \\[1.5mm]
       Is \textbf{explanation generation} from questions?  \\
       \> \cb\ Fully automated \cb\ Partially automated \cb\ Manual 
\end{tabbing}
\end{framed}
\caption{\label{fig:scoresheet}XAI scoresheet. Notation: alternatives (``pick one'') are separated by ``/'' whereas multiple options (``select all applicable'') are indicated with ``\cb''.} 
\end{figure}

\textbf{Global Explanations} capture what sort of information is available about the system's \emph{overall} functioning. One useful type of information is \emph{how} the system works. 
Another  is  \emph{how well} it works~\cite{DBLP:conf/chi/LiaoGM20,DBLP:journals/fcomp/HoffmanMKJT23}. This encompasses information on various limitations of the system (things it cannot do, including contexts in which it should not be used). It could also include information on the performance of the system (e.g.~how accurate is it, how reliable, and in what scope/context can this level of performance be expected). 
These two questions can be addressed by providing a static document, or an interactive manual that allows the stakeholder to gain understanding of respectively how and how well the system functions~\cite{IEEEP7001}. 
Additionally, for systems where data plays an important role in decision-making, part of the answers to ``how?'' and ``how well?'' is information about data used (e.g.~training data). This might usefully include the training data source, what steps were taken to ensure and/or assess its quality, information on distributions within the data (e.g.~breakdown by demographic factors), how it was processed, and what limitations or assumptions exist. For example, a data set of facial photos from a particular country reflects that country's demographics, and may not be appropriate to use in a country with significantly different demographics. 

\textbf{Local Explanations} are, unsurprisingly, a key part of the XAI scoresheet that capture a range of information. We begin with general information about the features of explanations that are generated. Firstly, since different people need different explanations, it can be useful to be able to generate different explanations for different people (``individually customised''). In order to do this it can be useful to be able to ``\ldots \textit{provide some information on what is desired in a good answer. For instance, how complete does the answer need to be? What is the aim of the person asking the question - are they a novice trying to clarify why something slightly unexpected occurred, i.e. to learn, or are they an expert seeking to dig deep to ascribe blame for something that should not have occurred?}''~\cite{WinikoffEMAS24}. 
Secondly, since explanations can be quite complex and large, it can be useful to make them \emph{interactive}~\cite{DBLP:journals/ki/HoffmanMKMC23}. For instance, provide a partial high-level answer to a question and allow the user to interactively get more information where needed. 
Finally, when an explanation is given, it can be useful for it to include indications of \emph{confidence}~\cite{WinikoffEMAS24}~(e.g.~that the system's decision was based on a particular belief that was held with a certain level of confidence), and of the \emph{scope}~\cite{DBLP:conf/fat/MittelstadtRW19}~(the extent to which the explanation generalises, e.g.~in a loan decision application that the key factor for a certain decision was the applicant's salary, but that this holds only as long as certain other factors are held within a certain range). 

Next the XAI scoresheet records information on what \emph{concepts} are used in explanations. This is useful to capture because it indicates at a high level what explanations look like. Furthermore, it has been argued~\cite{DBLP:journals/ai/WinikoffSDD21} that since humans explain their behaviour in terms of particular concepts such as beliefs, goals, and valuings~\cite{MalleBook}, using these same concept to explain autonomous systems can make explanations more accessible and understandable. The XAI scoresheet lists examples and features (since a range of explanation mechanisms use these), as well as a range of concepts for autonomous systems~(see~\cite[Chapter 2]{DBLP:books/daglib/0016041}) and values~\cite{DBLP:conf/aies/MehrotraJT21,DBLP:journals/tiis/MehrotraJJT24}.

Next, the scoresheet captures what sort of \emph{explanation types} the explanation generation system is able to generate. This is captured in terms of the sorts of questions that the system can answer. The most basic form of question is factual: e.g.~did something happen? We also consider the possibility of future-looking factual expectations: is something expected to happen? For example, a system that does some form of planning (reactive or first principles planning) may be able to provide information about what it did (in the past), and what it intends to do or expects to be the case (in the future). 

Perhaps the most common type of explanation considered in the literature is answering ``Why?'' questions. As for factual questions, explanations can potentially refer to both the past (e.g.~a certain course of action was selected because of past information or beliefs) and the future (e.g.~a certain action was performed in order to achieve a certain situation in the future). In addition to being able to pose ``Why?'' questions, it can also be useful to be able to ask ``Why not?'', and it has been argued~\cite{DBLP:journals/ai/Miller19} that as humans we naturally tend to ask \emph{contrastive} questions (``Why did you do $X$ rather than $Y$?'', although sometimes the ``rather than'' part is implicit). 

Finally, the literature identified a range of forms of \emph{hypothetical} question types that can be useful. For instance, Hoffman \textit{et al.}~\cite{DBLP:journals/ki/HoffmanMKMC23} note that contrastive and counterfactual explanations play a role in supporting a range of user goals.
Liao~\textit{et al.}~\cite{DBLP:conf/chi/LiaoGM20} identify a number of such question types: ``what-if?'' (what would happen in a different situation?), ``how to be?'' (how to change inputs to achieve a certain outcome), and ``how to still be?'' (what changes to inputs would leave the outcome unchanged).

The last part of the XAI scoresheet concerns automation. Ideally, when the user asks a question, the system generates the explanation. However, it is also possible to have the system support explanation construction by the user, or even provide enough information so the user can construct an explanation manually (e.g.~see \S\ref{sec:systemchimp}). 
However, a manual explanation construction process is clearly less desirable than having the system generate the explanation.

\subsection{Comparing with IEEE P7001}\label{sec:compare7001}

Having explained what we have included in our XAI scoresheet and why, we now briefly compare it to the IEEE P7001  transparency scoresheet~\cite{IEEEP7001}. Like us, IEEE P7001 proposed a scoresheet in order to help bridge the gap between high-level statements about desirable properties of systems and actionable metrics. However, there are a number of significant differences. The most significant difference is that P7001 is broader in scope, focusing on transparency, whereas we focus specifically on explainability. For example, P7001 includes requirements about warning bystanders that sensors are collecting information, and providing certification agencies with information about verification and validation activities that were done. Focusing on explainability aspects, P7001 is fairly limited, making our scoresheet useful and complementary. For instance, we also include information on veracity, on how \emph{well} the system works, and consider factors such as the level of confidence, scope of generalisation, concepts used, level of automation, and additional question types (``why not?'', contrastive questions, ``how to be?'', ``how to still be?'').

To illustrate, consider the example (Appendix B.2 of IEEE P7001) of a medical diagnosis AI. With respect to explainability (as opposed to transparency more broadly), the assessment in the appendix of IEEE P7001 specifies only that end users (i.e.~clinicians) need to be provided with (i) information on how the system functions (i.e.~global explanation) specifically in an interactive form, and (ii) with the ability to pose ``why?'' and ``what if?'' questions to the system (levels 1-4). Some things that are is missing from this assessment but captured by our XAI scoresheet are: 
\begin{itemize}
\item Veracity: how are explanations derived, and how can we know that an explanation corresponds to the actual reason?
\item Are explanations interactive? Do they they include indications of the system's confidence? Do they include an indication of the scope within which they are valid? 
\item Does the system support contrastive questions? Does it support other forms of hypothetical questions? (e.g.~``what would I need to change to get this (different) recommendation?'')
\end{itemize}
There are also some other differences including that IEEE P7001 gives a set of orthogonal transparency requirements, classified by stakeholder, whereas we do not classify by stakeholder, since there is not a clear distinction between stakeholders~\cite{DBLP:journals/fcomp/HoffmanMKJT23}; and that for incident investigators P7001 requires a blackbox (``Event Data Recorder''), whereas we do not require a blackbox, since it is not essential to providing explanations.

\section{Operationalising the scoresheet}\label{sec:operationalise}

In this section we consider the question of \emph{how} to use the scoresheet, in other words, when filling it out, how does one work out what the answers should be? We also note what other information is useful to capture (apart from what is in the scoresheet).

However, before starting to fill out the scoresheet for a given system that is being considered, we first need to identify who the relevant stakeholders are, and then what are their goals. We also need to identify for the application domain what are the risks that exist, and what level of risk is considered acceptable. This is required because to assess, for instance, whether there is adequate explanation of (globally) how the system operates, we are really answering the question of whether the provided information allows the stakeholders to gain an understanding of the system's functioning that is adequate for their goals. 
In other words, we need to know the stakeholders and their goals to assess this.  
For example, an elderly person using a domestic robot to support their independent living would need less information on how the robot functions and its limitations (e.g.~tasks it cannot do well) than an agency responsible for certifying these robots for domestic use. 
Similarly, in order to assess the system's reliability, we need to know what the needs are: what can go wrong, and what are the potential consequences? 

\textbf{Basic information:} this covers a few questions, that can be answered by asking the developer. However, although these questions appear to be answered by a simple ``yes'' or ``no'', they are actually an example of where there is  additional information that is not in the scoresheet itself that is useful to capture. For example, when indicating that there is access to the developers of the system, there is a range of other information that is important to consider and record. For instance: How accessible are the developers? How reliably and quickly are they likely to be able to respond to queries or to meeting requests? To what extent might developers be reluctant to be transparent, especially when doing so might reveal an area of weakness in the system's performance? Similar considerations apply to access to trusted domain experts. This additional information is not included in the scoresheet itself in order to keep the scoresheet a brief and useful summary. 

\textbf{Veracity:} If the system does not provide local explanations, then the answer to this is a simple ``not applicable''. Otherwise, to answer this question we need to consider the process by which (local) explanations are generated. The key question is: ``if I get an explanation, how confident can I be that this actually reflects the real reasons for the behaviour I am seeing?''. 
There are two approaches that can be used in order to complete this part of the scoresheet. Firstly, one can simply ask the system's developers to explain how explanations are generated (a meta-explanation), with particular emphasis on the links to the decision-making module and what steps were taken to ensure high reliability. Alternatively, it may be possible to evaluate veracity experimentally by setting up scenarios to see whether it is possible for explanations to deviate from the real reasons. For example, having (potentially adversarially-generated) scenarios $A$ and $B$ that give different behaviours but where the explanation provided for a question, such as ``Why did you do $A$?'' provides an explanation that only refers to features that are the same as in $B$.

\begin{figure}[t]
\begin{framed}
\begin{tabbing}
    XX \= XXX \= XXXXX \= XXX \= \kill
    \centerline{\textbf{Global Explanation} checklist} \\[1.5mm]
    There is an adequate description of: \\ 
    \cb\ \ldots \emph{how} the system operates, including  \\
    \> \cb\ \ldots  its (static) \emph{structure} \\
    \> \cb\ \ldots  its (dynamic) \emph{process} \\[1.5mm] 
    \cb\ \ldots \emph{how well} the system functions, including information on  \\
    \> \cb\ \ldots  the system's \emph{performance} \\
    \> \cb\ \ldots risks (including ethical issues) \\
    \> \cb\ \ldots  the system's \emph{limitations} \\
    \> \> (e.g.~situations in which it should (not) be used) \\[1.5mm]
    If the system uses training data: \\ 
        \> \cb\ Information about the training data is available \\ 
    \> \> (e.g.~its source, size) \\
    \> \cb\ \ldots including information on the process \\ 
    \> \> (e.g.~data selection, cleaning, etc.) 
\end{tabbing}
\end{framed}
\caption{Global Explanation Checklist\label{fig:checklist}} 
\end{figure}

\textbf{Global Explanations:} There are just two yes/no questions to be answered, but in order to answer them there is additional information that needs to be considered, and in fact we create an auxiliary checklist\footnote{We use the term ``checklist'' here since, unlike the scoresheet in Figure~\ref{fig:scoresheet}, all the responses here are ticks in boxes.} (Figure~\ref{fig:checklist}) 
to ensure that it is considered. In addition to the checklist, it can also be useful to capture in what form the global explanation is provided. For example, is it a static document or in an interactive form~\cite{IEEEP7001}? The explanation could also use a range of (possibly derived) models such as decision tree, rules, or weighted features~\cite{DBLP:conf/chi/LiaoGM20}.

The essential question here is whether information is provided (on ``how?'' and ``how well?'') in a form and at the level of detail that is appropriate for the relevant stakeholder(s), and whether the information provided is adequate for their needs. For example, the level of understanding of how a system operates may be lower for  a user of a system and higher for someone certifying the system for use in a given context. 

The two unindented checkboxes in Figure~\ref{fig:checklist} correspond to the two questions under Global Explanations in Figure~\ref{fig:scoresheet}. We would normally expect that in order to get an overall tick, the indented questions would also need to be ticked. 
It would clearly be unusual to indicate, for example, that there is an adequate description of how the system functions, without there being both descriptions of the system's (static) structure and its (dynamic) process of operation. 
Similarly, it would be unusual to consider information on how well the system functions to be adequate if it did not address the system's performance, the risks associated with its use, and its limitations. 

With regard to the questions under ``how well'', the first (``performance'') indicates whether information has been provided on how well the system operates within the intended domain of application. In other words, when the system is being used as intended, in a domain that it is designed for, how well does it perform, for instance, how accurate is it? 
The second (``risk'') indicates whether information has been provided on what risks exist in relation to the use of the system (including any ethical issues).
The third (``limitations'') indicates whether information has been provided on the boundaries of intended use: in what situations is the system's effectiveness reduced, or, indeed, the system should not be used? For example, an application for assessing loan applications may only be appropriate to use when the applicants are salaried employees. 
Edge cases can play a role in documenting these boundaries. 

If the system uses training data, then it can be important to also have information on the training data (such as where/how it was obtained, its size, and other characteristics such as demographic distributions), and on the process that was used to prepare the data (e.g.~selection, cleaning, quality assessment) and to use it (e.g.~training methodology, hyper-parameters).
There can also be relevant data-related information included in the discussion of limitations. For example, that a given data set only covers certain demographic groups adequately, so should not be used for other demographic groups. 
There is a range of work on how to provide information about data in this context that can be leveraged (e.g.~\cite{DBLP:journals/corr/abs-1808-07261,DBLP:journals/corr/abs-1805-03677,DBLP:journals/corr/abs-2201-03954,DBLP:journals/corr/abs-1803-09010,DBLP:journals/corr/abs-1810-03993}).

Finally, moving on to \textbf{Local Explanations}, recall that information in the scoresheet covers a range of things:  features of explanations, the concepts used, the explanation types supported, and the level of automation. 

The explanatory features provided (e.g.~individual customisation, interactivity) should be able to be determined by asking or just by using the system. For the first one (individual customisation), the question is whether it is possible for different people/roles asking the same question to be able to get different (relevant to each) answers. If the answer is yes, then it can be useful to also capture (not on the scoresheet) the \emph{extent} and \emph{forms} of individual customisation. For example, can a question include an explicit indication  of what level of detail is sought, or what concepts should be used? 

Identifying the concepts used in explanations requires looking at a range of explanations (and documentation). It may not always be immediately clear which parts of the explanation correspond to which concepts. For example, the explanation in Figure~\ref{fig:exampleexplanation} 
has a number of elements, and it may not be immediately clear which are beliefs, goals, or preferences. Identifying instances of concepts can be done by applying the definitions of the concepts (See Appendix A). 
For example, ``money was available'' is a factual statement about the environment, i.e.~a belief. On the other hand ``to allow you to catch the bus'' is a single desired state, i.e.~a goal, whereas a statement that compares more than one alternative indicates a preference (e.g. ``made choice \ldots has the shortest duration \ldots in comparison with \ldots''). Finally, if an explanation (or part of it) does not appear to map to any of the concepts, then it is an ``Other'' (e.g.~``I needed to buy a bus ticket in order to allow you to go by bus'' is an example of doing one thing in order to enable a later action).

\begin{figure}
\begin{framed}
``\textit{A bicycle was not available, money was available, the made choice (catch bus) has the shortest duration to get home (in comparison with walking) \ldots I needed to buy a bus ticket in order to allow you to go by bus, and I have the goal to allow you to catch the bus.}''
\end{framed}
\caption{Example Explanation from~\cite[\S2]{DBLP:conf/extraamas/WinikoffS23}\label{fig:exampleexplanation}}
\end{figure}

Similarly, identifying the forms of explanation types provided requires looking at a range of questions (and documentation), and may require some interpretation. For example a question of the form ``What situation would give an outcome of $X$?'' does not immediately correspond to the question types in the scoresheet. However, considering what is provided to the system (the desired behaviour) and what it provides to the human (the situation, i.e.~conditions under which the desired outcome occurs) can allow us to see that it corresponds to a form of ``How to be?'' - what situation will lead to desired behaviour (see also \S\ref{sec:marl}). 

Finally, identifying the level of automation should be straightforward.

\section{Applying to different use cases}\label{sec:apply}

In this section we demonstrate the utility and versatility of the scoresheet by applying it to a range of systems. This shows that it can be applied to a diverse range of systems, and also demonstrates that the scoresheet for a system summarises information about the explainability of the system in a useful form. 

We have selected the following six systems, which represent a broad range of types of intelligent or autonomous systems: 
(1) ChatGPT being used to recommend travel activities; 
(2) Generative AI being used to generate medical images; 
(3) A planner being used in a robotic application; 
(4) A search and rescue application implemented using BDI (Belief-Desire-Intention) concepts (goals, plans); 
(5) A multi-agent reinforcement learning system applied in a number of domains including a multi-robot search and rescue; and
(6) A taxi scheduling domain where the system combines learning and planning. See Figure \ref{fig:completedscoresheets} for the corresponding scoresheets.

We note that the scoresheets are based on the specific systems mentioned on an as-is basis, rather than what could be done to the systems to make them more explainable, as there are certainly ways to do so. 

\subsection{ChatGPT for activity recommendation}

We selected ChatGPT as an example of a general-purpose LLM, and applied it to the domain of generating recommendations for activities when visiting a city. The transcripts from our interaction with ChatGPT are available in Appendix B. 
In addition to  asking ChatGPT for recommendations, we also asked for a range of explanations. We were expecting ChatGPT to do relatively poorly, but in fact it did quite well in providing explanations (as indicated in the bottom of the scoresheet, see Figure~\ref{fig:completedscoresheets}).

However, it is important to note that there is no information on what measures (if any) have been taken to attempt to ensure that answers, including explanations, reflect the actual reasons. Since ChatGPT is known to bullshit~\cite{DBLP:journals/ethicsit/HicksHS24} (sometimes euphemistically termed ``hallucinate''), this is an issue, since it means that the explanations cannot be relied upon. This is highlighted in the scoresheet. 

\subsection{PET Image Generation}

This system uses generative AI to generate PET (Positron Emission Tomography) images ~\cite{DBLP:conf/dicta/MudiyanselageTTWCV23}. 
It takes PET images from one radiotracer and generates pseudo-PET images of another radiotracer. The training and test data were obtained from a hospital with appropriate privacy and ethics approvals. The scoresheet clearly captures that while there is information provided on both \emph{how} and \emph{how well} the system works, the system does not have the ability to explain specific images generated, other than providing a confidence level (e.g.~0.85).

\subsection{Planning for mobile service robot}\label{sec:systemchimp}

This system~\cite{servicerobotchimp} uses a hybrid planning system (CHIMP), that combines HTN-style task decomposition and meta-CSP search, resulting in an HTN planner able to handle very rich domain knowledge. This is applied to an application of a mobile service robot that performs tasks such as serving hot coffee with sugar. For such a task, it must reason not just about the consequences of each action but also the duration of the action, whilst considering  alternative possibilities for accomplishing the same task.

The system keeps a log of what was done and why. This makes it possible to obtain information to answer a broad range of questions. However, as highlighted in the scoresheet, this needs to be done manually by the developers. On the other hand, because this information is generated directly from the planner, the explanations can be relied upon.

\subsection{Search \& Rescue using BDI}

This system is a simulation that controls UAVs carrying out a search and rescue task \cite{rodriguez2023}. It is implemented using BDI concepts (goals and plans) in SARL~\cite{Rodriguez2014Sarl}, and uses the TriQPAN pattern~\cite{DBLP:conf/atal/SRJT24a,DBLP:conf/atal/SRJT24b} to extend SARL to be able to provide a range of (local) explanations. The scoresheet clearly indicates that the system is able to provide a range of explanations, and that this is fully automated. It also indicates that the explanations are directly derived from logs of the actual system, so the explanations can be relied upon.

\begin{figure*}[htp]
\includegraphics[width=0.9\textwidth]{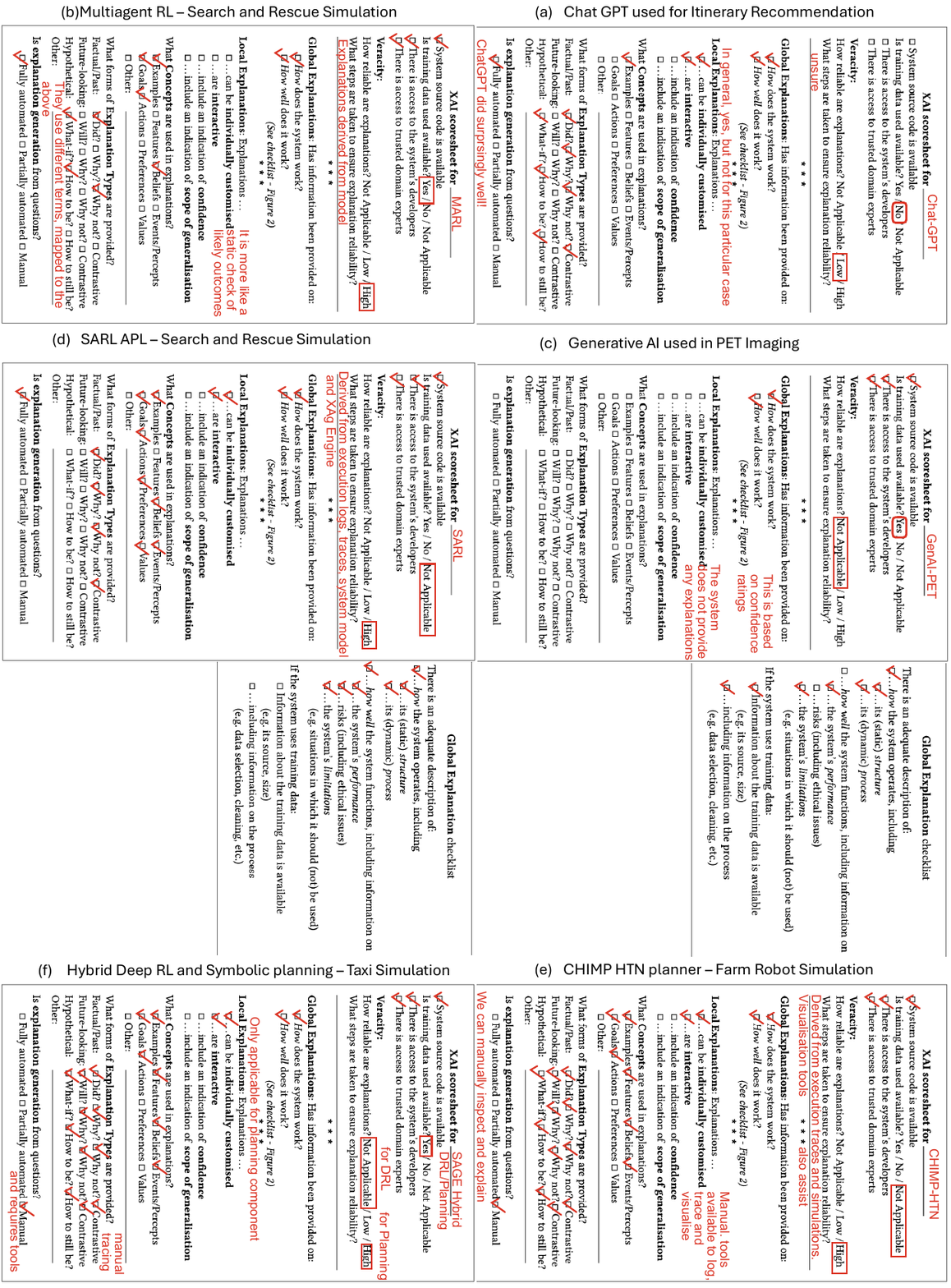} 
\caption{Completed Scoresheets for the Six Systems~\label{fig:completedscoresheets} }
\end{figure*}

\subsection{Multi-Agent Reinforcement Learning}\label{sec:marl}

This work extends multi-agent reinforcement learning with explanation features~\cite{DBLP:conf/ijcai/BoggessK022}, building on earlier work on single agent reinforcement learning explanation~\cite{DBLP:conf/hri/HayesS17}. They apply their approach to three domains: a multi-robot search and rescue scenario, a multi-robot cooperative delivery task, and a grid-based game where agents cooperate and compete to collect food. 

In essence, they provide two things: an algorithm to create a summary of a policy, and an algorithm to provide explanations for given queries (they extend this in a subsequent paper to temporal logic queries~\cite{DBLP:conf/ijcai/BoggessK023}).

The first contribution, a summary of a policy, is a \emph{global} explanation (``\textit{policy summarization provides a global view of the agent behavior under a MARL policy}''~\cite[\S4]{DBLP:conf/ijcai/BoggessK022}). However, while the query-based explanations provide what look like typical local explanations, in fact the explanations are in terms of \emph{likely paths}, rather than in terms of a particular execution of the system. 

Regardless of this though, it is interesting to observe that the three question types they support do not match in an obvious way to the question types that we have included in our scoresheet. Specifically, the first question type (``When do [agents] do [actions]?'') is used ``\textit{for identifying conditions for action(s) of a single or multiple agent(s)}''~\cite[\S4]{DBLP:conf/ijcai/BoggessK022}. This can be seen, in intent, if not phrasing, as being related to ``how to be?'': it is identifying conditions that allow particular actions (i.e.~behaviours) to occur. The second question type (``Why don't [agents] do [actions] in [states]?'') is clearer, corresponding to our ``Why not?''. Finally, the third question type that they support (``What do [agents] do in [predicates]?'') is used ``\textit{for revealing agent behavior under specific conditions}'' (ibid) and can be seen as a form of ``what if?'': given particular conditions, what would happen?

The scoresheet clearly captures that this system provides local explanations of various types, and that the explanation generation is done directly from the behaviour-generating module, and hence the explanations can be relied upon.

\subsection{Taxi planning using learning \& planning}

This work~\cite{DBLP:conf/ausai/ChesterDZT23} proposes an architecture that combines planning and learning, and demonstrates it in a taxi planning domain. The architecture has three levels: a top-level that uses reinforcement learning to identify what are the best goals to select, a middle level that uses an off-the-shelf planner to develop plans to achieve these goals, and a low-level module that uses deep reinforcement learning to perform low-level actions within the plans. 

In terms of using the scoresheet to assess the explainability aspects of this system a key challenge is that it has three modules, each of which has different explainability features. The planning module (similar to \S\ref{sec:systemchimp}) captures information that can be used to (manually) generate (highly reliable) explanations. However, the deep reinforcement learning module does not provide any form of explainability. 

There are two ways in which this can be captured using the scoresheet. The first (which is preferred) is to use a single scoresheet for the whole system, but annotate it to indicate when answers apply to only parts of the system. For example, for veracity we might indicate that it is ``Not Applicable'' for the RL part of the system and ``High'' for the planning component. 
The second way, which may be required if the first approach yields an overly cluttered and complex scoresheet, is to have a separate scoresheet for different modules in the system (perhaps with a system-wide scoresheet that refers to them).

\section{Discussion \& Conclusion}\label{sec:closing}

\balance 

We have presented a scoresheet for explainability, along with detailed guidance for how to use it. The scoresheet was then applied to a broad range of systems, demonstrating its usability and generality.
Looking at the results of applying the scoresheet (Figure~\ref{fig:completedscoresheets}) we can see that important explainability features of the different systems are captured. 
For example, for ChatGPT it is clear that explanations may not be reliable, but that the system provides a range of explanation types. On the other hand, for PET image generation, the scoresheet captures clearly that only global explanations are available.  For the mobile service robot  the scoresheet clearly indicates that a range of (local) explanations are available, and that they can be relied upon (because they are generated directly from the planner), but that the construction of explanations from the information is a manual process. The search and rescue (using SARL) and Multi-agent reinforcement learning are similar in providing a range of (reliable) explanations, and do not require manual construction of these explanations. Finally, the taxi planning application scoresheet captures clearly that there are multiple modules in the system, and that these have different explainability characteristics.

\subsection{Limitations \& Future Work}
One limitation is that the scoresheet has only been used by the authors. Therefore, future work includes further use and evaluation of the scoresheet. This could include having a range of people (e.g.~ various roles, covering the stakeholder types discussed in \S\ref{sec:relatedwork}, as well as a range of experience levels and diverse demographics) use it to assess systems. It could also include assessing how well the scoresheet can be used for other use cases (e.g.~specifying the explainability requirements of an application, rather than assessing a given system). 
This would be done by indicating what XAI features are required of a system that is to be used in a certain context, e.g.~if a bank was looking to develop a system for making loan decisions it could use the scoresheet to specify what XAI features would be required for the system-to-be.
Indeed, it might be possible to use a scoresheet to specify the explainability requirements for a whole sector or domain (e.g.~transport, policing), or even to specify regulatory requirements relating to explainability. 

Finally, we highlight some broader research challenges for the XAI research community. 
There is a need to move beyond explaining particular decisions or actions (local explanations) to be able to provide useful information on \emph{how} the system works, using local explanations to illustrate (i.e.~``global-local'' explanations), including highlighting edge cases~\cite{DBLP:journals/fcomp/HoffmanMKJT23}. 
There is also a need to be able to identify and include information in particular about behaviours that are \emph{surprising}~\cite{DBLP:journals/ki/HoffmanMKMC23}.

\begin{acks}
This research is partially supported by the C2IMPRESS project funded by the EU.
\end{acks}



\clearpage 



\newpage 

\appendix



\section*{Supplementary material}

# A  Concept Definitions

An **Example** is a particular instance of the system at hand being used: what was the input and what was the output or outcome.

A **Feature** is an "individual measurable property or characteristic of a phenomenon" [Bis07].

A **Belief** is "some aspect of the agent's knowledge or information about the environment, itself or other agents" [PW04, §2.3].

An **Event** is a "significant occurrence that the agent should respond to in some way" [PW04, §2.2].

A **Percept** is "an item of information received from the environment by some sensor" [PW04, §2.1].

A **Goal** is "something the agent is working on or towards" [PW04, §2.2].

An **Action** is something that an agent does and which affects its environment.

A **Preference** is an indication that one (or more) actions or outcomes are preferred to another set of (one or more) actions or outcomes.

**Values** are universal high-level drivers of human behaviour [Sch12], for example, seeking power over others.

**Valuings** are things that "directly indicate the positive or negative affect toward the action or its outcome" [Mal04, Section 4.2.4]. For example, someone appreciating snow.

# B ChatGPT Transcripts

**You said:** give me a 3 day itinerary for wellington NZ

**ChatGPT said:** Certainly! Wellington, New Zealand, offers a rich blend of cultural experiences, outdoor adventures, and culinary delights. Here's a suggested 3-day itinerary to make the most of your time in this vibrant city:

**Day 1: Explore Wellington City**

**Morning:**

1. **Te Papa Tongarewa:** Start your day with a visit to New Zealand's national museum. Explore exhibits on Maori culture, natural history, and contemporary art.
2. **Waterfront Promenade:** Take a leisurely stroll along the waterfront promenade, enjoying views of the harbor and city skyline.
3. **Cable Car and Botanic Garden:** Ride the historic Wellington Cable Car to the Wellington Botanic Garden. Spend some time wandering through the beautiful gardens and enjoying panoramic views of the city.

**Lunch:** 4. **Cuba Street:** Head to Cuba Street, Wellington's most famous culinary and cultural precinct. Choose from a range of cafes and restaurants offering diverse cuisines.

**Afternoon:** 5. **Weta Workshop:** Visit the Weta Workshop, famous for its work on the Lord of the Rings and The Hobbit films. Take a guided tour to see how movie props and special effects are created. 6. **Mount Victoria Lookout:** Drive or hike up to Mount Victoria Lookout for stunning panoramic views of Wellington and its surroundings. 7. **Wellington Museum:** Dive into Wellington's history at the Wellington Museum, housed in the historic Bond Store building on the waterfront.

**Dinner:** 8. **Dine in Courtenay Place:** Courtenay Place is the heart of Wellington's dining and nightlife scene. Choose from a range of restaurants offering local and international cuisine.

**Day 2: Nature and Wildlife**

**Morning: Zealandia Ecosanctuary:** Spend the morning at Zealandia, an urban ecosanctuary renowned for its native birdlife and lush forest trails. Take a guided tour or explore on your own.

**Lunch:** 2. **Picnic at Oriental Bay:** Grab some takeaway food and head to Oriental Bay for a picnic on the sandy beach. Alternatively, dine at one of the waterfront cafes.

**Afternoon:** 3. **Seal Coast Safari:** Take an afternoon tour to the rugged south coast with Seal Coast Safari. Marvel at the dramatic coastline, spot seals basking on the rocks, and learn about

the area's natural history. 4. **Red Rocks Reserve:** After the safari, explore Red Rocks Reserve, known for its unique geological formations and seal colony.

**Dinner:** 5. **Seafood Feast:** Treat yourself to a seafood dinner at one of Wellington's top seafood restaurants, offering fresh local catches.

**Day 3: Day Trip and Cultural Immersion**

**Morning: Wellington Cable Car Museum:** Start your day with a visit to the Wellington Cable Car Museum, located at the cable car's Lambton Quay terminus. Learn about the history of this iconic Wellington attraction.

**Day Trip:** 2. **Wairarapa Wine Region:** Take a day trip to the nearby Wairarapa wine region. Explore vineyards, taste award-winning wines, and enjoy a leisurely lunch at a winery restaurant.

**Afternoon:** 3. **Maori Cultural Experience:** Return to Wellington in the afternoon and immerse yourself in Maori culture. Attend a cultural performance, visit a marae (Maori meeting house), or join a guided tour to learn about Maori history and traditions.

**Evening:** 4. **Farewell Dinner:** Enjoy a farewell dinner at one of Wellington's top restaurants, savoring delicious cuisine paired with local wines or craft beers.

This itinerary offers a mix of Wellington's top attractions, outdoor adventures, cultural experiences, and culinary delights, providing a comprehensive taste of what the city and its surroundings have to offer. Adjustments can be made based on your interests and preferences.

**You said:** why did you recommend Wairarapa wine region?

**ChatGPT said:**

The Wairarapa wine region is recommended for a day trip from Wellington because it offers a fantastic opportunity to explore New Zealand's renowned wine industry in a picturesque setting. Here are a few reasons why it's worth considering:

1. **Proximity:** The Wairarapa wine region is located just a short drive from Wellington, making it easily accessible for a day trip. It's about an hour's drive from Wellington, which allows for a convenient visit without requiring an overnight stay.

2. **Scenic Beauty:** The Wairarapa region is known for its stunning landscapes, including rolling vineyards, lush countryside, and charming rural towns. It offers a refreshing change of scenery from the urban environment of Wellington, providing a relaxing and enjoyable day out.

3. **Wine Tasting:** The Wairarapa is renowned for its boutique wineries, particularly those in the subregions of Martinborough, Gladstone, and Masterton. Visitors can explore a variety of cellar doors, tasting rooms, and vineyard cafes, sampling premium wines such as Pinot Noir, Sauvignon Blanc, and Riesling.

4. **Culinary Experiences:** In addition to wine tasting, many wineries in the Wairarapa region offer gourmet food experiences, such as platters of local cheeses, artisanal bread, and other delicacies. It's an opportunity to indulge in delicious food while enjoying scenic views of the vineyards.

5. **Culture and History:** The Wairarapa region has a rich history of winemaking dating back to the late 19th century. Visitors can learn about the area's viticultural heritage, explore historic vineyards and wineries, and gain insights into the winemaking process through guided tours and tastings.

Overall, a day trip to the Wairarapa wine region provides a perfect blend of wine tasting, culinary delights, scenic beauty, and cultural experiences, making it a memorable addition to your Wellington itinerary.

**You said:** why didn't you recommend kafe eis?

**ChatGPT said:** Kafe Eis is a popular spot in Wellington known for its gelato and desserts, and it's definitely worth considering for a visit, especially if you have a sweet tooth or enjoy indulging in delicious treats. Here are a few reasons why Kafe Eis could be a great addition to your Wellington itinerary:

1. **Delicious Gelato:** Kafe Eis is renowned for its high-quality gelato made from fresh, locally sourced ingredients. With a wide variety of flavors ranging from classic favorites to unique creations, there's something to satisfy every palate.

2. **Dessert Options:** In addition to gelato, Kafe Eis offers a tempting selection of desserts, including cakes, pastries, and sweet treats. Whether you're craving a decadent slice of cake or a refreshing sorbet, you'll find plenty of options to choose from.

3. **Café Atmosphere:** Kafe Eis provides a cozy and welcoming atmosphere, making it a great place to relax and enjoy a sweet treat with friends or family. Whether you choose to sit inside the café or take your gelato to go, you can soak in the vibrant ambiance of Wellington's bustling Cuba Street precinct.

4. **Convenient Location:** Located in the heart of Wellington's Cuba Street, Kafe Eis is easily accessible and conveniently situated near other popular attractions, shops, and restaurants. It's a convenient stop to make while exploring the city center.

While Kafe Eis wasn't specifically mentioned in the itinerary, it could certainly be included as a delightful stop for dessert or a refreshing gelato break during your time in Wellington. Feel free to incorporate it into your plans based on your interests and cravings!

**You said:** can you recommend me an itinerary for wellington NZ for one day. I do not have a car

**ChatGPT said:** Absolutely! Here's a one-day itinerary for Wellington, New Zealand, tailored for travelers without access to a car:

**Morning:**

**1. Te Papa Tongarewa - National Museum of New Zealand:** Start your day with a visit to Te Papa Tongarewa, New Zealand's national museum. Explore its diverse exhibits on Maori culture, natural history, art, and New Zealand's social and cultural history. Entry is free, although donations are appreciated.

**2. Wellington Waterfront:** Take a leisurely stroll along the Wellington Waterfront, enjoying the views of the harbor, sculptures, and vibrant atmosphere. You can stop for a coffee at one of the waterfront cafes and watch the boats go by.

**Midday:**

**3. Cable Car and Botanic Garden:** Catch the iconic Wellington Cable Car from Lambton Quay to the Wellington Botanic Garden. Enjoy a scenic ride up the hillside and explore the beautiful gardens. Don't miss the stunning views of the city from the Cable Car Lookout.

**4. Lunch at Zealandia:** Head to the Zealandia Ecosanctuary for lunch at their cafe. Enjoy a meal surrounded by native bush and birdlife. Zealandia offers a unique opportunity to see some of New Zealand's rarest wildlife in a protected environment.

**Afternoon:**

**5. Weta Workshop:** Take a guided tour of the Weta Workshop, renowned for its work on films like The Lord of the Rings and Avatar. Learn about the movie-making process and see props, costumes, and special effects up close.

**6. Oriental Bay:** Spend some time relaxing at Oriental Bay, Wellington's most popular inner-city beach. Take a stroll along the promenade or dip your toes in the water if the weather is warm.

**Evening:**

**7. Cuba Street:** Explore Cuba Street, Wellington's vibrant cultural hub. Wander down the pedestrianized street lined with shops, cafes, restaurants, and street performers. Stop for dinner at one of the many eateries offering diverse cuisines.

**8. Night Market (if available):** If you're visiting on a Friday evening (between October and April), check out the Wellington Night Market on Cuba Street. Sample a variety of international street food, enjoy live music, and soak up the lively atmosphere.

This itinerary allows you to experience some of Wellington's top attractions, scenic spots, and

cultural highlights, all within walking distance or easily accessible by public transportation. Enjoy your day in Wellington!

**You said:** why did you not recommend wairarapa wine region

**ChatGPT said:** In this one-day itinerary for Wellington without access to a car, I didn't include a visit to the Wairarapa wine region for a few reasons:

1. **Distance:** The Wairarapa wine region is located about an hour's drive from Wellington, making it less feasible for a day trip without access to a car. While there are tour options available, they may require a significant portion of your day for transportation alone, limiting the time you have to explore other attractions in Wellington.

2. **Time Constraints:** A day trip to the Wairarapa wine region would likely require a substantial time commitment, leaving less time to explore Wellington itself. Given the limited timeframe of one day in the city, I prioritized attractions and experiences that are easily accessible and can be enjoyed within walking distance or short public transportation rides.

3. **Focus on Wellington:** Since the itinerary is specifically for exploring Wellington, I aimed to include activities and attractions within the city that showcase its culture, history, and natural beauty. While the Wairarapa wine region is undoubtedly a fantastic destination for wine enthusiasts, it's a separate experience that would warrant its own day trip with more time dedicated to exploring the vineyards and cellar doors.

However, if you have more time available during your visit to Wellington or if you're particularly interested in wine tasting, you could certainly consider extending your stay in the region or joining a guided tour that includes transportation from Wellington to the Wairarapa.

**You said:** why did you recommend Cable car instead of tour of parliament

**ChatGPT said:** Including the Cable Car and Botanic Garden in the itinerary instead of a tour of the Parliament building was a choice based on offering a variety of experiences that showcase different aspects of Wellington. Here are some reasons for that decision:

1. **Iconic Experience:** The Wellington Cable Car is an iconic attraction in the city, offering visitors not only a practical mode of transportation but also stunning views of the cityscape. It's a quintessential Wellington experience that many visitors enjoy.

2. **Scenic Beauty:** The Cable Car journey takes you from the heart of the city up to the Wellington Botanic Garden, offering a scenic ride through picturesque surroundings. Once at the top, visitors can explore the beautiful gardens, enjoy panoramic views of the city, and relax in a tranquil setting.

3. **Accessibility:** The Cable Car is easily accessible from the city center, making it a convenient option for visitors without access to a car. It's also suitable for all ages and fitness levels, making it inclusive for a wide range of travelers.
   4. **Limited Time:** With only one day to explore Wellington, the itinerary aimed to include attractions that are easily manageable within a short timeframe. The Cable Car and Botanic Garden offer a compact yet rewarding experience that can be enjoyed in a relatively short amount of time, leaving room for other activities.

While a tour of the Parliament building is undoubtedly an interesting and informative experience, it may require more time and advance planning, including booking a tour slot and going through security checks. Depending on individual interests and preferences, visitors to Wellington may choose to prioritize either the Cable Car and Botanic Garden or a tour of the Parliament building based on their available time and preferences.

**You said:** what should i do if i also wanted to do a cultura maori night

**ChatGPT said:** If you're keen to include a cultural Māori night experience in your one-day Wellington itinerary, you could make a few adjustments to fit it in. Here's a revised version of the itinerary that incorporates a Māori cultural evening:

**Morning:**

### 1. Te Papa Tongarewa - National Museum of New Zealand:

Start your day with a visit to **Te Papa**, particularly focusing on the **Mana Whenua** exhibit, which highlights Māori culture, history, and traditions. This will give you an excellent introduction to Māori heritage before your evening experience.

### 2. Wellington Waterfront:

Stroll along the waterfront to enjoy the harbor views and sculptures. You can grab a coffee at a café here and take in the city's atmosphere.

**Midday:**

### 3. Wellington Cable Car and Botanic Garden:

Take the Wellington Cable Car to the Botanic Garden. Walk through the gardens and enjoy the panoramic city views. You could explore the Carter Observatory or simply enjoy a leisurely walk among the flowers.

### 4. Lunch in Cuba Street:

Head to **Cuba Street** for lunch. It's full of vibrant cafes and restaurants with various international cuisines to choose from. Pick a spot that suits your taste and soak in the lively atmosphere.
**Afternoon:**

### 5. Parliament Tour (Optional):

If you want to add a touch of history and politics, you could replace or combine this with a brief **Parliament tour**. The Beehive and Parliament House offer free guided tours that give you insights into New Zealand's government and political history.

### 6. Oriental Bay:

After lunch, take a walk or relax at **Oriental Bay**. It's a beautiful urban beach where you can unwind, take in the sea air, or even dip your toes in the water.

**Evening:**

### 7. Māori Cultural Experience:

For an authentic **Māori cultural evening**, you can book a performance and dinner experience. While Wellington doesn't have as many Māori cultural shows as other parts of New Zealand (such as Rotorua), some tour companies and venues do offer cultural performances, traditional waiata (songs), haka (ceremonial dance), and hangi (a traditional Māori feast).

**Options for a Māori Cultural Experience:**

- **Te Wharewaka o Poneke:** Located along the waterfront, this venue often hosts Māori cultural experiences, including traditional performances, storytelling, and occasionally a hangi meal.
- **Pataka Art + Museum** in nearby Porirua sometimes hosts Māori cultural events, but you'd need to check their schedule ahead of time.

You can also inquire with your accommodation or Wellington i-SITE visitor information center about any available Māori cultural nights during your visit.

**Alternative Cultural Evening:**

If a full Māori cultural night isn't available, you can still enjoy a cultural-themed dinner at restaurants offering modern Kiwi cuisine with Māori and Pacific Island influences. For example, **HIAKAI** in Wellington offers dishes inspired by Māori traditions and ingredients.

By including a cultural Māori night, you get a well-rounded experience of Wellington, combining the city's urban attractions, natural beauty, and New Zealand's indigenous culture.

**You said:** if i make it half day can i still take the cable car

**ChatGPT said:** Yes, you can definitely take the Cable Car even if you're planning a half-day itinerary in Wellington! The Wellington Cable Car is a quick and iconic experience, and it's easily manageable within a shorter timeframe. Here's how you can incorporate it into a half-day itinerary:

**Half-Day Wellington Itinerary (No Car Needed)**

**Morning:**

**1. Te Papa Tongarewa - National Museum of New Zealand (1-2 hours):**

Start your half-day at **Te Papa** to explore key exhibits on New Zealand's history, culture, and Māori heritage. Focus on the main exhibits or visit the **Mana Whenua** section for a quick dive into Māori culture. **2. Cable Car Ride & Botanic Garden (1-1.5 hours):**

- Walk to the **Wellington Cable Car** station at Lambton Quay (a short 15-minute walk from Te Papa).
- Take the Cable Car up to the Botanic Garden. Enjoy the panoramic views at the top, stroll through the gardens, or visit the Cable Car Museum. The ride itself only takes about 5 minutes each way, so you can explore as much or as little of the gardens as your time allows.

**3. Quick Lunch (Optional, 30 minutes):**

Head to **Cuba Street** for a quick lunch or coffee before heading to your next destination. Cuba Street is full of fast options, including cafes, bakeries, and casual eateries.

**If Time Allows:**

If you're doing this in the afternoon or have extra time, you can also fit in a walk along the **Wellington Waterfront** or visit **Oriental Bay** for a relaxing view of the harbor.

The Cable Car and Botanic Garden are flexible activities that can easily fit into a half-day visit to Wellington. You'll still have time to explore some of the city's key attractions while enjoying a mix of cultural and scenic experiences.



\end{document}